\documentclass[10pt,twocolumn,letterpaper]{article}

\usepackage[pagenumbers]{cvpr} 

\usepackage{graphicx}
\usepackage{amsmath}
\usepackage{amssymb}
\usepackage{booktabs}

\usepackage[dvipsnames]{xcolor}
\usepackage{listings}
\usepackage{enumitem}

\usepackage[normalem]{ulem}
\usepackage{bm}
\usepackage{multirow}

\usepackage{algorithm}
\usepackage{algorithmic}

\usepackage[pagebackref,breaklinks,colorlinks]{hyperref}

\DeclareMathOperator*{\argmax}{argmax}

\usepackage[capitalize]{cleveref}
\crefname{section}{Sec.}{Secs.}
\Crefname{section}{Section}{Sections}
\Crefname{table}{Table}{Tables}
\crefname{table}{Tab.}{Tabs.}

\def\cvprPaperID{} 
\def\confName{CVPR}
\def\confYear{2022}

\begin{document}

\title{A Fast Knowledge Distillation Framework for Visual Recognition}

\author{Zhiqiang Shen~~~~~Eric Xing\\
$^{}$~~~~CMU ~\&~ $^{}$MBZUAI\\
{\tt\small zhiqians@andrew.cmu.edu~~epxing@cs.cmu.edu}
}
\maketitle

\begin{abstract}
While Knowledge Distillation (KD) has been recognized as a useful tool in many visual tasks, such as supervised classification and self-supervised representation learning, the main drawback of a vanilla KD framework is its mechanism, which consumes the majority of the computational overhead on forwarding through the giant teacher networks, making the entire learning procedure inefficient and costly.
ReLabel~\cite{yun2021re}, a recently proposed solution, suggests creating a label map for the entire image. During training, it receives the cropped region-level label by RoI aligning on a pre-generated entire label map, allowing for efficient supervision generation without having to pass through the teachers many times. However, as the KD teachers are from conventional multi-crop training, there are various mismatches between the global label-map and region-level label in this technique, resulting in performance deterioration. In this study, we present a Fast Knowledge Distillation (FKD) framework that replicates the distillation training phase and generates soft labels using the multi-crop KD approach, while training faster than ReLabel since no post-processes such as RoI align and softmax operations are used. When conducting multi-crop in the same image for data loading, our FKD is even more efficient than the traditional image classification framework. On ImageNet-1K, we obtain 79.8\% with ResNet-50, outperforming ReLabel by $\sim$1.0\% while being faster. On the self-supervised learning task, we also show that FKD has an efficiency advantage. Our project page is \href{http://zhiqiangshen.com/projects/FKD/index.html}{here}, source code and models are available at: \url{https://github.com/szq0214/FKD}. 
\end{abstract}

\vspace{-0.15in}
\section{Introduction}
\label{sec:intro}

Knowledge Distillation (KD)\cite{hinton2015distilling} has been a widely used technique in various visual domains, such as the supervised recognition~\cite{romero2014fitnets,yim2017gift,xie2020self,muller2019does,shen2021label,beyer2021knowledge} and self-supervised representation learning~\cite{shen2021s2,fang2021seed,caron2021emerging}. The mechanism of knowledge distillation is to force the student to imitate the output of a teacher network or ensemble teachers, as well as converging on the ground-truth labels. Given the parameters $\bf \theta$ of the target student at iteration $(t)$, we can learn the next iteration parameters $\theta^{(t+1)}$ by minimizing the following objective which contains two terms:
\vspace{-0.1in}
\begin{equation} \label{kd_math}
\begin{array}{c}
\vspace{0.05in} 
\boldsymbol{\theta}_\text{student}^{(t+1)}=\arg \min _{\theta \in \Theta} \frac{1}{N} \sum_{n=1}^{N}(1-\lambda) \mathcal{H}\left(\boldsymbol{y}_{n}, \boldsymbol{S}_{\theta}\left(\boldsymbol{x}_{n}\right)\right) \\ \vspace{-0.1in} \quad+\lambda \mathcal{H}\left(\boldsymbol{T}^{(t)}(\boldsymbol{x}_{n}), \boldsymbol{S}_{\theta}\left(\boldsymbol{x}_{n}\right)\right) 
\end{array}
\end{equation}
where $\boldsymbol{y}_{n}$ is the ground-truth label for $n$-th sample. $\boldsymbol{T}^{(t)}$ is the teacher's output at iteration $(t)$ and $\boldsymbol{S}_{\theta}(\boldsymbol{x}_{n})$ is the student's prediction for the input sample $\boldsymbol{x}_{n}$. $\mathcal{H}$ is the cross-entropy loss function. $\lambda$ is coefficient for balancing the two objectives. The first term aims to minimize the entropy between one-hot ground-truth label and student's prediction while the second term is to minimize between teacher and student's predictions. The teacher $\boldsymbol T$ can be pre-trained in either a supervised or self-supervised manner. Many literature~\cite{shen2020meal,yun2021re,beyer2021knowledge,shen2021label} have empirically shown that the first term of true hard label in Eq.~\ref{kd_math} is not required on larger-scale datasets like ImageNet~\cite{deng2009imagenet} under the circumstance that the teacher or ensembled teachers are accurate enough. In this work, we simply minimize the soft predictions between teacher and student models for the fast distillation design.

\begin{table}[t]
\centering
\caption{A feature-by-feature comparison between ReLabel~\cite{yun2021re} and our FKD framework on various elements and properties.}
\label{tab:my-table_dataset}
\vspace{-0.1in}
\resizebox{0.48\textwidth}{!}{
\begin{tabular}{l|c|c|c|c}
\toprule[1.1pt]
 Method & \bf Generating label   & \bf Label storage & \bf Info. loss & \bf Training \\ \midrule 
Vanilla KD      &  Implicit    &  None & \color{green}{No}  & \color{yellow}{Slow} \\  
ReLabel~\cite{yun2021re} &  Fast  &  Efficient  & \color{red}{Yes}  & \color{red}{Fast} \\  
FKD (Ours)      &  Slow   & Efficient & \color{green}{No} & \color{green}{Faster} \\
\bottomrule[1.1pt]
\end{tabular}
}
\vspace{-0.1in}
\end{table}

\begin{figure*}[t]
  \centering
  \includegraphics[width=0.86\textwidth]{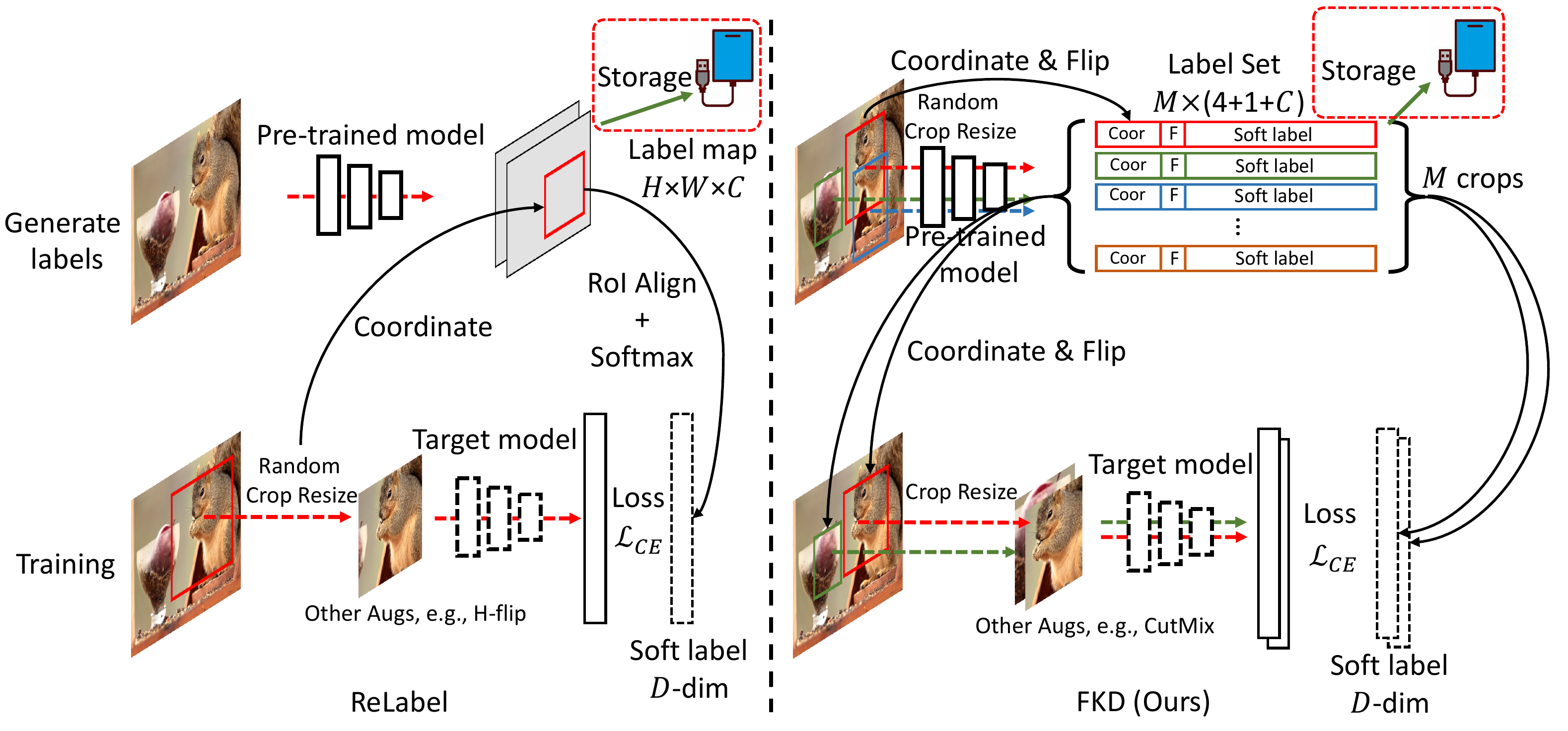}
  \vspace{-0.12in}
  \caption{Comparison of ReLabel~\cite{yun2021re} and our fast knowledge distillation (FKD) framework.} 
  \label{fig:dis_com}
  \vspace{-0.13in}
\end{figure*}

The fundamental disadvantage in such a paradigm, according to KD's definition, is that a considerable proportion of computing resources is consumed on passing training data through large teacher networks to produce the supervision $\boldsymbol{T}^{(t)}$ in each iteration, rather than updating or training the target student parameters. Intuitively, the forward propagation through teachers can be shared across epochs since the parameters of them are frozen for the entire training. Considering that the vanilla distillation framework itself is basically inefficient, how to reduce or share the forward computing of teacher networks across different epochs becomes the core for accelerating  KD frameworks. 
A natural solution to overcome this drawback is to generate one probability vector as the soft label of input data corresponding to each image in advance, and then reuse them circularly for different training epochs. However, in modern network training, we usually impose various data augmentation strategies, particularly the random crop technique, which causes the inconsistency in which the simple global-level soft vector for the entire image can no longer  accurately reflect the true probability distribution of the local input region after these augmentations.

To address the data augmentation, specially random-crop caused inconsistency issue in generating one global vector to the region-level input, while preserving the advantage of soft label property, ReLabel~\cite{yun2021re} is proposed to store the global label map annotations from a pre-trained strong teacher for reutilization by RoI align~\cite{he2017mask} without passing through the teacher networks repeatedly. Fig.~\ref{fig:dis_com} (left) shows a full explanation of this mechanism. However, due to the different input processes on teachers, this strategy is essentially not equivalent to the vanilla KD procedure. The mismatches are primarily due to two factors: {\bf(i)} the teacher network is usually trained with a random-crop-resize scheme, whereas in ReLabel, the global label map is obtained by feeding into the global image, which cannot exactly reflect the soft distribution as distillation process whose random-crop-resize operation is employed in the input space; {\bf(ii)} RoI align cannot guarantee the distribution is completely identical to that from the teachers' forwarding.

In this work, we introduce a Fast Knowledge Distillation (FKD) framework to overcome the mismatching drawback and further avoid information loss on soft labels. Our strategy is straightforward: As shown in Fig~\ref{fig:dis_com} (right), in the label generation phase, we directly store the soft probability from multiple random-crops into the label files, together with the coordinates and other data augmentation status like flipping. During training, we assign these stored coordinates back to the input image to generate the crop-resized input for passing through the networks, and computing the loss with the corresponding soft labels. The advantages of such a strategy are two folders: {\bf(i)} Our region-based generating process and obtained soft label for each input region are identical to the vanilla KD's output, implying that no information is lost during the label creation phase; {\bf(ii)} Our training phase enjoys a faster pace since no post-process is required, such as RoI align, softmax, etc. We can further assign multiple regions from the same image in a {\em mini}-batch to facilitate the burden of data loading.

We demonstrate the advantages of our FKD in terms of accuracy and training speed on supervised and self-supervised learning tasks. In the supervised learning scheme, we compare the baseline ReLabel and vanilla KD (Oracle) from scratch across a variety of backbone network architectures, such as CNNs, vision transformers, and the competitive MEAL V2 framework with pre-trained initialization. Our FKD is $\sim$1\% higher and slightly faster than ReLabel on ImageNet-1K, and 3$\sim$5$\times$ faster than oracle KD with similar performance. On the self-supervised learning manner, we employ S$^2$-BNN as the baseline for verifying the speed advantage of our proposed framework.

Our contributions of this work:
\vspace{-0.08in}
\begin{itemize}[leftmargin=0.14in]
	\addtolength{\itemsep}{-0.08in}
	\item We present a fast knowledge distillation (FKD) framework that achieves the same high level of performance as vanilla KD, while keeping the same training speed and efficiency as non-KD training without information loss.
	\item We reveal a discovery that in image classification frameworks, one image can be sampled with multiple crops within a {\em mini}-batch to facilitate data loading and speed up training, meanwhile without sacrificing performance.
	\item To prove the effectiveness and versatility of our approach, we demonstrate FKD on a variety of tasks and distillation frameworks, including supervised classification and self-supervised representation learning.
\end{itemize}

\section{Related Work}

\noindent{\textbf{Knowledge Distillation.}} The principle behind Knowledge Distillation~\cite{hinton2015distilling} is that a student is encouraged to emulate or mimic the teachers' prediction, which helps the student generalize better on unseen data. One core advantage of distillation is that the teacher can provide softened distribution which contains richer information about the input data compared to the traditional one-hot labels, especially when the data augmentation such as random cropping is used on the input space. Distillation can avoid incorrect labels by predicting them from the strong teachers in each iteration, which reflects the real situation of the transformed input data. Conventionally, we can impose a temperature on the logits to re-scale the output distributions from teacher and student models to amplify the inter-class relationship on supervisions and allow for improved distillation. Recently, many variants and extensions are proposed~\cite{papernot2016distillation,huang2017like,wang2018dataset,zhang2019your,muller2019does,park2019relational,shen2021label,xie2020self,yin2020dreaming,chung2020feature,stanton2021does}, such as employing internal feature representations~\cite{romero2014fitnets}, adversarial training with discriminators~\cite{shen2019meal}, transfer flow~\cite{yim2017gift}, contrastive distillation~\cite{tian2019contrastive}, patient and consistent~\cite{beyer2021knowledge} etc. For the broader overviews of related methods for knowledge distillation, please refer to~\cite{gou2021knowledge,9340578}.

\noindent{\textbf{Efficient Knowledge Distillation.}} Improving training efficiency for knowledge distillation is crucial for pushing this technique to a wider usage scope in real-world applications. Previous efforts on this direction are generally not sufficient. ReLabel~\cite{yun2021re} is a recently proposed solution that addresses this inefficient issue of KD surpassingly. In particular, it generates the global label map for the strong teacher and then reuses them through RoI align across different epochs. Our proposed FKD approach in this paper lies in an essentially different consideration and solution. We consider the property of vanilla KD to generate the randomly cropped region-level soft labels from the strong teachers and store them in advance, then reuse them by allocating them to different epochs in training. Our approach enjoys the same accuracy as vanilla KD and same training speed as regular non-KD classification frameworks, making it superior than ReLabel in both performance and training speed. 

\vspace{-0.06in}
\section{Approach} \label{approach_ana}
\vspace{-0.03in}

In this section, we begin by introducing some observations and properties on ReLabel's global-level soft label and FKD's region-level soft label distributions. Then, we present the detailed workflow of our FKD framework and elaborately discuss the generated label quality, training speed and the applicability on supervised and self-supervised learning. Finally, we analyze the strategies of label compression and storage for practical usage.

\begin{figure}[t]
  \centering
  \includegraphics[width=0.47\textwidth]{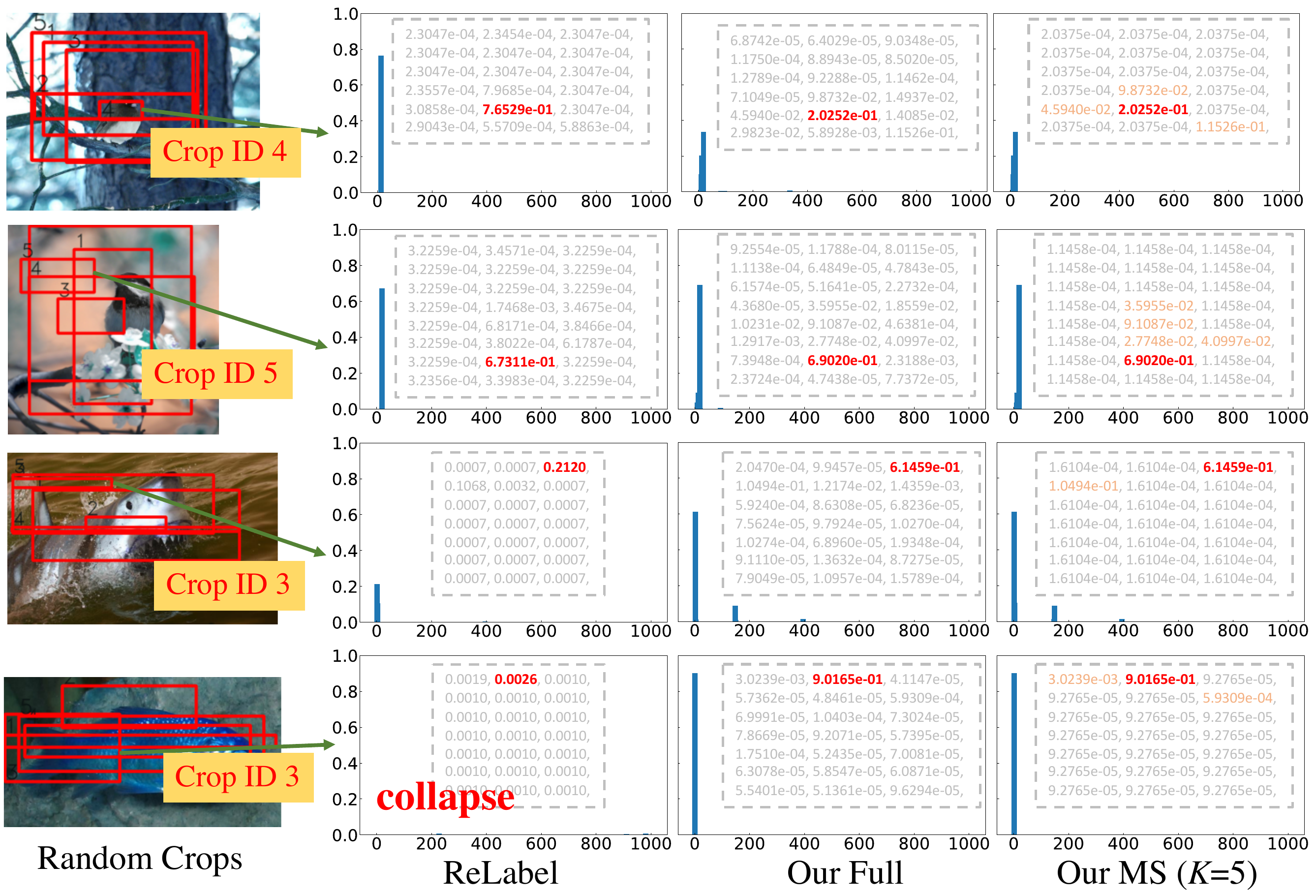}
  \vspace{-0.13in}
  \caption{Illustration of label distributions of ReLabel~\cite{yun2021re}, our FKD full label and our quantized label (Top-$5$). ``MS'' denotes the marginal smoothed labels, more details can be referred in Sec.~\ref{storage_s}. Grey numbers in each block are the corresponding partial probabilities/labels (limited by space) from different frameworks.} 
  \label{fig:dis_label_comparison}
  \vspace{-0.14in}
\end{figure}

\noindent{\textbf{Preliminaries: Limitations of Previous Solution}}\\
The mechanism of ReLabel through RoI align operation (an approximation solution) is naturally different from the vanilla KD when generating region-level soft labels. In Fig.~\ref{fig:dis_label_comparison}, we visualize the region-level label distributions of ReLabel and FKD on ImageNet-1K, and several empirical observations are noticed: {\bf(i)} ReLabel is more confident in many cases of the regions, so the soft information is weaker than our FKD. We conjecture this is because ReLabel feeds the global images into the network instead of local regions, which makes the generated global label map encode more global category information and ignores the backgrounds, as shown in Fig.~\ref{fig:dis_label_comparison} (row 1). Though sometimes the maximal probabilities are similar between ReLabel and FKD, FKD still contains more informative subordinate probabilities in the label distribution, as shown in Fig.~\ref{fig:dis_label_comparison} (row 2); {\bf(ii)} for some outlier regions, our strategy is substantially more robust than ReLabel, such as the loose bounding boxes of objects, partial object, etc., as shown in Fig.~\ref{fig:dis_label_comparison} (row 3);  {\bf(iii)} In some particular circumstance, ReLabel is unexpectedly collapsed with nearly uniform distribution, while our FKD still works well, as shown in the bottom row of Fig.~\ref{fig:dis_label_comparison}.

Moreover, there are existing mismatches between the soft label from ReLabel and oracle teacher prediction in KD when employing more data augmentations such as Flip, Color jittering, etc., since these augmentations are randomly applied during training. In ReLabel design, we cannot take them into account and prepare in advance when generating the global label map. In contrast, FKD is adequate to handle this situation: it is effortless to involve extra augmentations and record all information (ratio, degree, coefficient, etc.) for individual region from same or different images, and generate corresponding soft label by feeding the transformed image regions into the pre-trained teacher networks.

\begin{figure*}[h]
  \centering
  \includegraphics[width=0.75\textwidth]{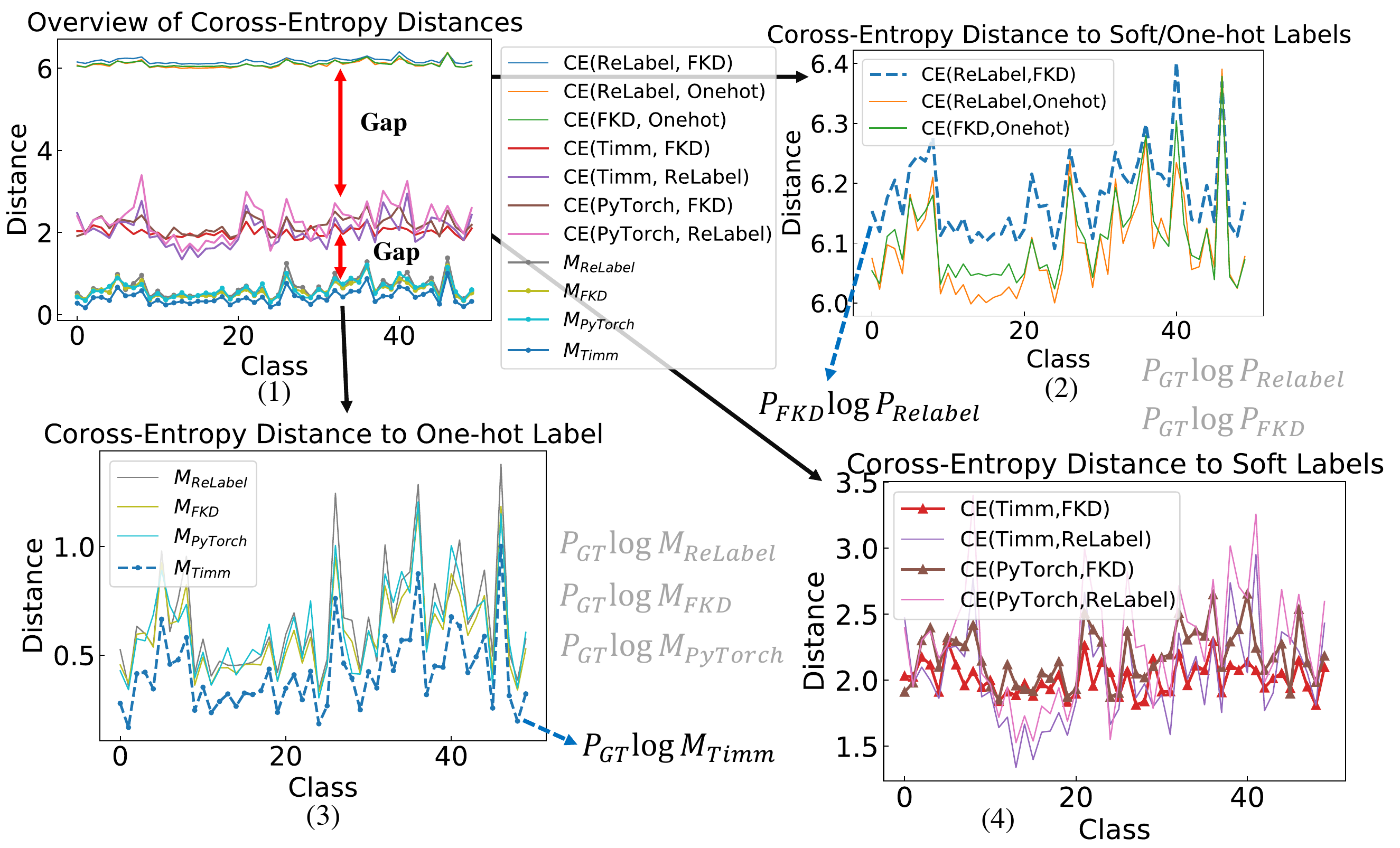}
  \vspace{-0.15in}
  \caption{Entropy distance analysis between different pairs of soft/one-hot labels and different labels trained model predictions. (1) is the overall distance visualization. (2), (3), (4) represent each detailed group in (1). We illustrate the first 50 classes in ImageNet-1K dataset.} 
  \label{fig:CE_analysis}
  \vspace{-0.13in}
\end{figure*}

\subsection{Fast Knowledge Distillation}
\vspace{-0.05in}

In a conventional visual training system, the bottleneck is usually from the network passing and data loader, while in a distillation framework, besides these computational consumptions, giant teachers have been the biggest burden for training. Our FKD aims to solve this intractable drawback.

\noindent{\textbf{Label Generation Phase.}} Following the regular random-crop resize training strategy, we randomly crop $\bm M$ regions from one image and employ other augmentations like flipping on them, then input these regions into the teachers to generate the corresponding soft label vectors $\bm P_i$, i.e., $\bm P_i=\boldsymbol T(\bm R_i)$ where $\bm R$ is the transformed region by transformations $ \mathcal{F}$ and $\boldsymbol T$ is the pre-trained teacher network, $i$ is the region index. We store all the region coordinates and augmentation hyper-parameters $\mathcal{F}$ with the soft label $\bm P$ for the following training phase, as shown in Fig.~\ref{fig:dis_com} (upper right). A detailed analysis of how to store these required values is provided in the following section.

\noindent{\textbf{Training Phase.}} In the training stage, instead of randomly generating crops as the conventional image classification strategy, we directly load the label file, and assign our stored crop coordinates and data augmentations for this particular image to prepare the transformed region-level inputs. The corresponding soft label will be used as the supervision of these regions for training. With the cross-entropy loss, the objective is: $\mathcal{L} =-\sum_i \bm P_i {\bf \log}\boldsymbol{S}_{\theta}(\boldsymbol{R}_{i})$, where $\boldsymbol{S}_{\theta}(\boldsymbol{R}_{i})$ is the student's prediction for the input region $\boldsymbol{R}_{i}$,  $\theta$ is the parameter of the student model that we need to learn. The detailed training procedure is shown in Fig.~\ref{fig:dis_com} (bottom right).

\subsection{Higher Label Quality}

\noindent{\textbf{Distance Analysis.}} We analyze the quality of various formulations of labels through the entropy distance with measures on their mutual cross-entropy matrix. We consider three types of labels: (1) human-annotated one-hot label, ReLabel, and our FKD. We also calculate the distance on the predictions of four pre-trained models with different accuracies, including: PyTorch pre-trained model (weakest), Timm pre-trained model~\cite{wightman2021resnet} (strongest), ReLabel trained model and FKD trained model. An overview of our illustration is shown in Fig.~\ref{fig:CE_analysis}. The upper curves, as well in (2), are averaged cross-entropy across 50 classes of (ReLabel$\rightarrow$FKD), (ReLabel$\rightarrow$One-hot) and (FKD$\rightarrow$One-hot). Here, we derive an important observation:
\begin{equation}
    (\mathcal{D}^{CE}_{R\rightarrow F} =-\bm P_{FKD} {\bf \log}\bm P_{ReLabel}) > (\mathcal{D}^{CE}_{R\rightarrow O} \ \ \bm {OR} \ \ \mathcal{D}^{CE}_{F\rightarrow O})
\end{equation}
where $\mathcal{D}^{CE}_{R\rightarrow F}$ is the  cross-entropy value of ReLabel $\rightarrow$ FKD. Essentially, FKD soft label can be regarded as the oracle KD label and $\mathcal{D}^{CE}_{R\rightarrow F}$ is the distance to such ``KD ground truth''. From Fig.~\ref{fig:CE_analysis} (2) we can see its distance is even large than ReLabel and FKD to the one-hot label. Since ReLabel (global-map soft label) and FKD (region-level soft label) are greatly discrepant from the one-hot hard label, the gap between ReLabel and FKD (``KD ground truth'') is fairly significant and considerable. If we shift attention to the curves of $\mathcal{D}^{CE}_{R\rightarrow O}$ and $\mathcal{D}^{CE}_{F\rightarrow O}$, they are highly aligned across different classes with similar values. In some particular classes, $\mathcal{D}^{CE}_{F\rightarrow O}$ are slightly larger. This is sensible as one-hot label is basically not the ``optimal label'' we desired.

In the bottom group, i.e., Fig.~\ref{fig:CE_analysis} (3), the entropy values are comparatively small. This is because they are from the pre-trained models and they have the decent performance under the criterion metric of one-hot label. Among them, $M_{Timm}$ has the minimal cross-entropy to the one-hot label, this is expected since the timm model is optimized thoroughly to fit the one-hot label with the highest accuracy.

\begin{figure}[t]
  \centering
  \includegraphics[width=0.48\textwidth]{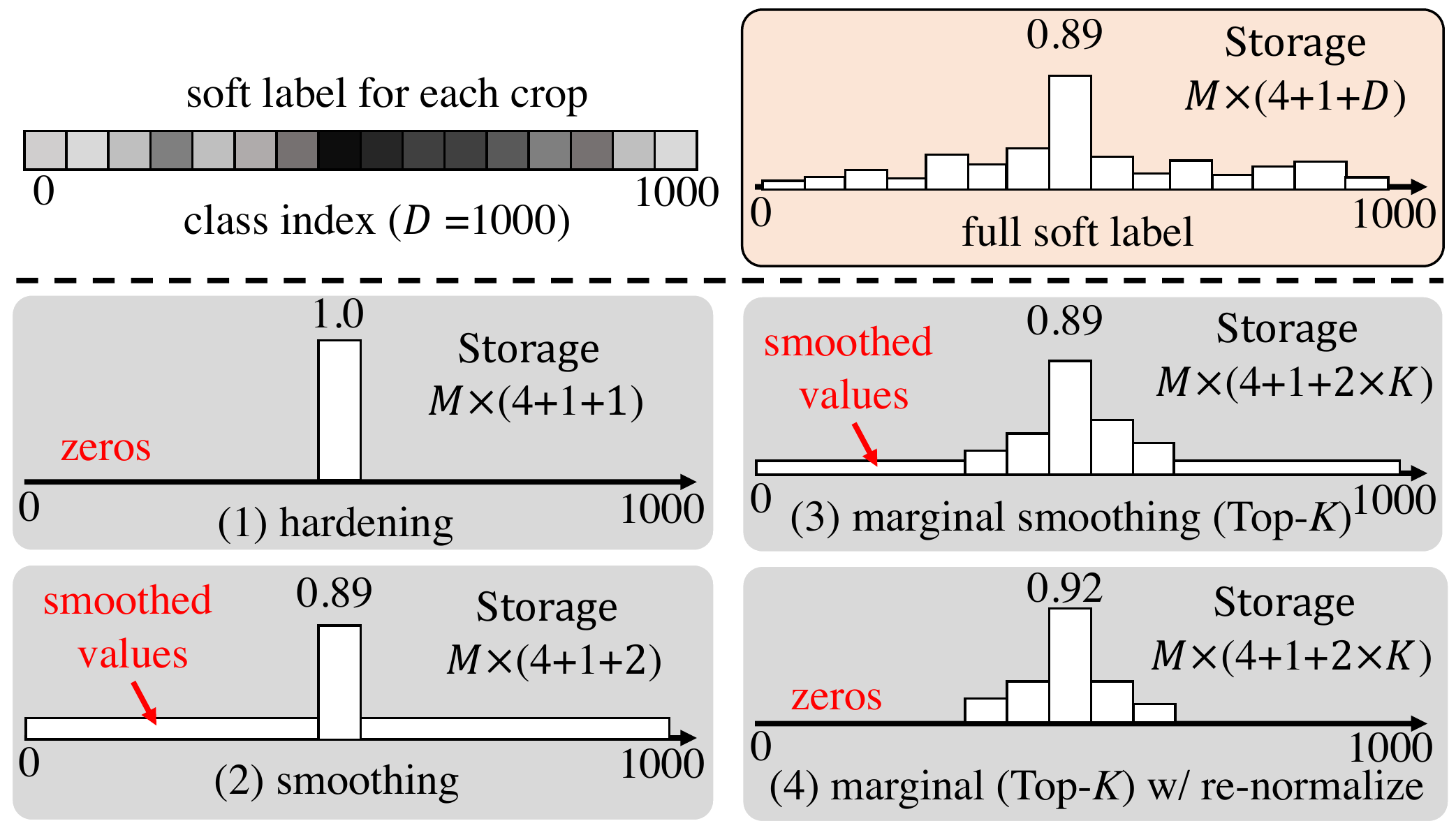}
  \vspace{-0.25in}
  \caption{Different label compression strategies and storage analysis for our fast knowledge distillation (FKD) framework.} 
  \label{fig:dis_label_compress}
  \vspace{-0.15in}
\end{figure}

In Fig.~\ref{fig:CE_analysis} (4), $\mathcal{D}^{CE}_{Timm\rightarrow F}$ and $\mathcal{D}^{CE}_{PT\rightarrow F}$ lie in the middle of $\mathcal{D}^{CE}_{Timm\rightarrow R}$ and $\mathcal{D}^{CE}_{PT\rightarrow R}$ with smaller variances. This reflects that FKD is more stable to the pre-trained models.

\subsection{Faster Training Speed} \label{FTS}
\noindent{\textbf{Multi-crop sampling within a {\em mini}-batch}.} As illustrated in Fig.~\ref{fig:dis_com} (right), we can use multiple crops in the same image to facilitate loading image and label files. Intuitively, it will reduce the diversity of training samples in a {\em mini}-batch since some of samples are from the same image. However, our experimental results indicate that it will not hurt the model performance, in contrast, it even boosts the accuracy when the number of crops from the same image is within a reasonable range (e.g., 4$\sim$8). 

\noindent{\textbf{Serrated learning rate scheduler.}} Since FKD samples multiple crops ($\#crop$) from one image, when iterating over the entire dataset once, we actually train the dataset $\#crop$ epochs with the same learning rate. It has no effect while using milestone/step {\em lr} scheduler, but it will change the {\em lr} curve to be serrated if applying continuous {\em cosine} or {\em linear} learning rate strategies. This is also the potential reason that multi-crop training can improve the accuracy.

\noindent{\textbf{Training Time Analysis: 1. Data Load}}

Data loading strategy in FKD is efficient. For instance, when training with a {\em mini}-batch of 256, traditional image classification framework requires to load 256 images and ReLabel will load 256 images + 256 label files, while in our method, FKD only needs to load $\frac{256}{\#crop}$ images + $\frac{256}{\#crop}$ label files, even faster than traditional training if we choose a slightly larger value for $\#crop$ (when $\#crop>$2)\footnote{We assume that loading each image and label file will consume the similar time by CPUs.}.

\textbf{2. Label Preparation}

\begin{figure}[t]
  \centering
  \includegraphics[width=0.48\textwidth]{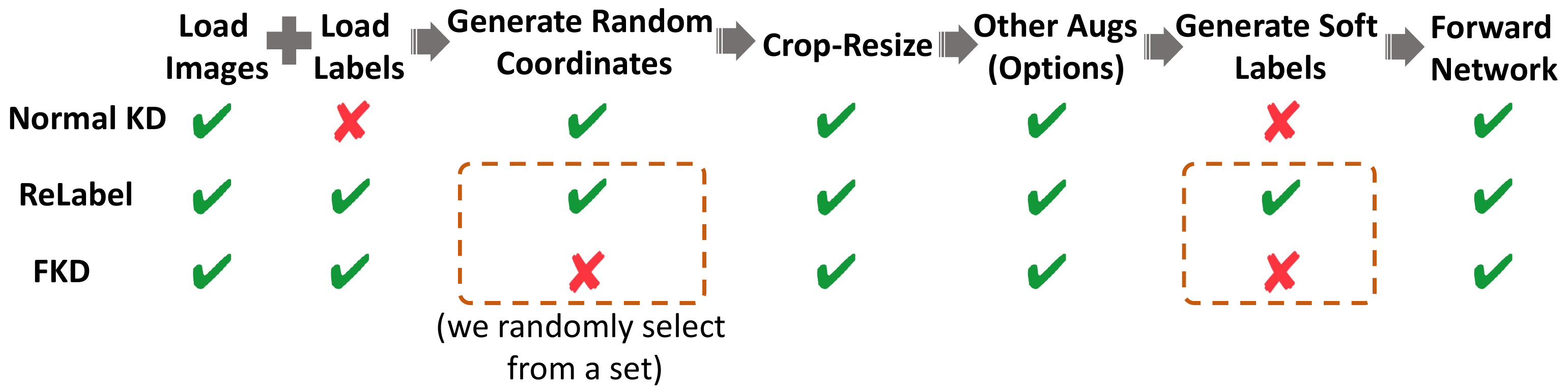}
  \vspace{-0.3in}
  \caption{Training flow analysis for normal KD, ReLabel~\cite{yun2021re} and our fast knowledge distillation (FKD) framework. {\color{Maroon} Maroon} dashed boxes indicate that the processes are required by ReLabel only while not existing in our FKD. Note that ``generate soft labels'' indicates RoI align + softmax in ReLabel. We both have the recovering process from the compressed label to full soft label as discussed in Sec.~\ref{FTS}.} 
  \label{fig:time_analysis}
  \vspace{-0.18in}
\end{figure}

We assign \#crop regions in an image to the current {\em mini}-batch for training. Since we store the label probability after {\em softmax} (in supervised learning), we can use assigned soft labels for the {\em mini}-batch samples directly without any post-process. This assignment is fast and efficient in implementation with a {\em randperm} function in PyTorch~\cite{paszke2019pytorch}. If the label is compressed using the following strategies, we will operate with an additional simple recovering process (as shown in Fig.~\ref{fig:dis_label_compress}) to obtain $D$-way soft label distributions. Note that ReLabel also has this process so the time consumption on this part will be similar to ReLabel. A detailed workflow and item-by-item comparison is shown in Fig.~\ref{fig:time_analysis}.

\begin{table*}[t]
\centering
\caption{A detailed comparison of different label quantization/compression strategies on ImageNet-1K. $M$ is the number of crops within an image and here we choose 200 crops as an example to calculate the space consumption. $N_\text{image}$ is the number of images, i.e., 1.2M for ImageNet-1K. $S_\text{LM}$ is the size of label map. $C_\text{class}$ is the number of classes. $D_\text{DA}$ is the parameter dimension of data augmentations to store.}
\label{tab:my-table_true_store}
\vspace{-0.1in}
\resizebox{0.995\textwidth}{!}{
\begin{tabular}{l|c|c||c|c|c|c|c|c}
\toprule[1.pt]
  &  ReLabel (Full)~\cite{yun2021re}&  ReLabel (Top-$5$)~\cite{yun2021re} &  Full   &   Hard &  Smoothing &  M Re-Norm ($K$=$5$) &  MS ($K$=$5$)  &  MS ($K$=$10$)\\ \midrule 
  Calculation & $N_\text{image}\!\times \! S_\text{LM}\! \times \! C_\text{class}$ & $N_\text{image}\!\times \! S_\text{LM}\! \times \! 2C_\text{Top-5}$ &   $N_\text{image}\!\times \! (C_\text{class}\!+\!D_\text{DA})$ & $N_\text{image}\!\times \! (1\!+\!D_\text{DA})$ &  $N_\text{image}\!\times \! (2\!+\!D_\text{DA})$   & $N_\text{image}\!\times \! (2K\!+\!D_\text{DA})$  &  $N_\text{image}\!\times \! (2K\!+\!D_\text{DA})$ & $N_\text{image}\!\times \! (2K\!+\!D_\text{DA})$  \\  
  Dim. of Soft Label & $15\times15\times1,000$ & $15\times15\times10$ & $M\times$1,000  & $M\times$1   & $M\times$2  &  $M\times$10  & $M\times$10 & $M\times$20   \\  
  + Coordinate \& Flip & -- & -- & $M\times$1,005  & $M\times$6   & $M\times$7  &  $M\times$15  & $M\times$15 & $M\times$25   \\  
 Real Cons. on Disk & $\sim$1TB & 10GB & $\sim$0.9TB  &  5.3GB  & 6.2GB & 13.3GB & 13.3GB & 22.2GB\\
\bottomrule[1.pt]
\end{tabular}
}
\vspace{-0.16in}
\end{table*}

\subsection{Training Self-supervised Model with Supervised Scheme}
\vspace{-0.05in}

In this section, we introduce how to apply our FKD for extending to the self-supervised learning (SSL) with faster training speed, comparing to the widely-used Siamese SSL frameworks. The label generation (from the self-supervised strong teachers), label preparation and training procedure are similar to the supervised scheme. However, we {\em keep the projection head in original SSL teachers and store the soft labels before softmax} for operating temperature\footnote{The temperature $\tau$ is applied on the {\em logits} before the {\em softmax} operation for self-supervised distillation.}.

\subsection{Label Compression and Storage Analysis} \label{storage_s}
We consider and introduce the following four strategies for compressing soft label for storage, an elaborated comparison of them can be referred to Table~\ref{tab:my-table_true_store}.
\vspace{-0.12in}

\begin{itemize}[leftmargin=0.1in]
	\addtolength{\itemsep}{-0.08in}
	\item \textbf{Hardening.} In hardening quantization strategy, the hard label $\bm Y_{\mathrm{\bm H}}$ is generated using the index of the maximum logits from the teacher predictions of regions. In general, label hardening is the one-hot label with correction by strong teacher models in region-level space.
	\vspace{-0.05in}
	\begin{equation} \label{hardening}
    \bm Y_{\mathrm{H}}=\argmax_{\bm c} \bm z_{\mathrm{FKD}}(\bm c)
    \vspace{-0.05in}
	\end{equation}
	where $\bm z_{\mathrm{FKD}}$ is the logits for each randomly cropped region produced by our FKD process.
	\item \textbf{Smoothing.}  Smoothing quantization replaces one-hot hard label $\bm Y_{\mathrm{\bm H}}$ with a mixture of soft $\bm y_{\bm c}$ and a uniform distribution same as label smoothing~\cite{szegedy2016rethinking}:
	\vspace{-0.05in}
\begin{equation}
\bm y^{\bf S}_{\bm c}=\left\{\begin{array}{ll}\bm p_{\bm c} & \text { if } \bm c=\text{\em hardening\ \ label}, \\ (1-\bm p_{\bm c}) / (\bm C-1) & \text { otherwise. }\end{array}\right.
\end{equation}
where $\bm p_{\bm c}$ is the probability after {\em softmax} at $c$-th class and $\bm C$ is the number of total classes. $(1-\bm p_{\bm c}) / (\bm C-1)$ a small value for flattening the one-hot labels. ${\bm y^{\bf S}_{\bm c}} \in {\bm Y_{\bf S}}$ is the smoothed label at $\bm c$-th class.

	\item \textbf{Marginal Smoothing with Top-$K$ (\em MS).} Marginal smoothing quantization reserves more soft information (Top-$K$) of teacher prediction than the single smoothing label $\bm Y_{\mathrm{\bm S}}$:
    \begin{equation}
    {\bm y}^{\bf MS}_{\bm c}=\left\{\begin{array}{ll}{\bm p}_{\bm c}  & \text{ if }  {\bm c}\in \{{\bf{Top-}}{K}\}, \\ \\  \frac{1-\sum\limits_{\bm c\in \{{\bf{Top-}}{K}\}} \bm p_{\bm c}} {\bm C-K}& \text { otherwise. }\end{array}\right.
    \end{equation}
where ${\bm y^{\bf MS}_{\bm c}} \in {\bm Y_{\bf MS}}$ is the marginally smoothed label at $\bm c$-th class. 

	\item \textbf{Marginal Re-Norm with Top-$K$ (\em MR).} Marginal re-normalization will re-normalize Top-$K$ predictions to $\sum_{\bm c\in \{{\bf{Top-}}{K}\}} \bm p_{\bm c}\!=\!1$ and maintain other logits to be zero (this strategy is spiritually similar to ReLabel~\cite{yun2021re} but slightly different in implementation as its input is logits before softmax so it used {\em softmax} while we use {\em normalize}, resulting in that our values outside  Top-$K$ remain zero.): 
	\vspace{-0.07in}
	\begin{equation}
    \bm y^{\bf M}_{\bm c}=\left\{\begin{array}{ll}\bm p_{\bm c}  & \text { if }  \bm c\in \{{\bf{Top-}}{K}\},  \\  0 & \text { otherwise. }\end{array}\right.
    \end{equation}
    \begin{equation}
    \bm y^{\bf MR}_{\bm c}=\text{Normalize}(\bm y^{\bf M}_{\bm c})=\frac{\bm y^{\bf M}_{\bm c}}{\sum_{\bm c=1}^C(\bm y^{\bf M}_{\bm c})}
    \end{equation}
where ${\bm y^{\bf MR}_{\bm c}} \!\!\in\!\! {\bm Y_{\bf MR}}$ is the re-normalized label at $\bm c$-th class.
\end{itemize}

\section{Experiments}

\noindent{\textbf{Experimental Settings and Datasets.}} Detailed lists of our hyper-parameter choices are shown in Table~\ref{tab:my-table_relabel_ours_com}, Table~\ref{tab:my-table_distillation} of Appendix. For the transparency and reproducibility of our framework, unless noted otherwise, we did not involve extra data augmentations (beyond the basic random crop and random horizontal flip) such as RandomAug~\cite{cubuk2020randaugment}, MixUp~\cite{zhang2018mixup}, CutMix~\cite{yun2019cutmix}, etc., in all of our experiments. Except for experiments on MEAL V2, we use {\em EfficientNet-L2-ns-475}~\cite{tan2019efficientnet,xie2020self} as the teacher model, we also tried weaker teacher but the performance in our experiment is slightly worse. For MEAL V2, we follow its original design by using {\em SENet154 + ResNet152\_v1s} (gluon version) ensemble as the soft label.

ImageNet-1K~\cite{deng2009imagenet} is used for the supervised classification and self-supervised representation learning. COCO~\cite{lin2014microsoft} is used for the transfer learning experiments in this work.

\noindent{\textbf{Network Architectures.}} Experiments are conducted on Convolutional Neural Networks~\cite{lecun1995convolutional}, such as ResNet~\cite{he2016deep}, MobileNet~\cite{howard2017mobilenets} and Vision Transformers~\cite{vaswani2017attention,dosovitskiy2020image}, such as DeiT~\cite{touvron2021training}, SReT~\cite{shen2021sliced}. For binary backbone, we use ReActNet~\cite{liu2020reactnet} in the self-supervised experiments.

\noindent{\textbf{Learning Schemes.}} We consider three training manners in vision tasks: (i) conventional supervised training from scratch; (ii) supervised fine-tuning from pre-trained parameters; and (iii) self-supervised distillation from scratch.

\begin{table}[t]
\centering
\caption{Comparison between ReLabel~\cite{yun2021re} and our FKD on ImageNet-1K. ``$\Diamond$'' denotes our training following the same protocol in Table~\ref{tab:my-table_relabel_ours_com} w/o distillation. Models are trained from scratch.}
\label{tab:my-table_relabel}
\vspace{-0.1in}
\resizebox{0.49\textwidth}{!}{
\begin{tabular}{l|c|c|c|c}
\toprule[1.pt]
 Method &  Network   &  Top-1 (\%) &  Top-5 (\%) &  Training Time \\ \midrule 
Vanilla$\Diamond$      &  ResNet-50    &  78.1  & 94.0 & 1.0 \\ 
ReLabel~\cite{yun2021re} &  ResNet-50  &  78.9  & --  &  ~~~~~~~$\uparrow$0.5\%~\cite{yun2021re} \\  
FKD (Ours)      &  ResNet-50   &  ~~~~~~~\bf 79.8$^{\bf{+0.9}}$ &  \bf 94.6 & \bf $\downarrow$0.5\% \\ \hline
Vanilla$\Diamond$      &  ResNet-101    & 79.7  &  94.6 & 1.0 \\  
ReLabel~\cite{yun2021re} &  ResNet-101  &  80.7  & --  &  ~~~~~~~$\uparrow$0.5\%~\cite{yun2021re} \\  
FKD (Ours)      &  ResNet-101   &  ~~~~~~~\bf 81.7$^{\bf{+1.0}}$ & \bf 95.6 & \bf $\downarrow$0.5\% \\ 
\bottomrule[1.pt]
\end{tabular}
}
\vspace{-0.1in}
\end{table}

\begin{table}[t]
\centering
\caption{Comparison of MEAL V2~\cite{shen2020meal} and our FKD on ImageNet-1K. ``w/ FKD'' denotes the model is trained using the same protocol as original MEAL V2, i.e., all the same hyper-parameters. ``$\heartsuit$'' represents the training using {\em cosine lr} and $1.5\times$ epochs. Models are trained from the pre-trained initialization.} 
\label{tab:my-table_MEALV2}
\vspace{-0.1in}
\resizebox{0.49\textwidth}{!}{
\begin{tabular}{l|c|c|c|c|c}
\toprule[1.pt]
 Method &  Network  &  \#Params  &  Top-1  &  Top-5 &  Speedup \\ \midrule 
MEAL V2~\cite{shen2020meal} &  ResNet-50  & 25.6M &  80.67  & 95.09  &  1.0  \\  
MEAL V2 w/ FKD &  ResNet-50   & 25.6M & 80.70 &  95.13 & \bf ~~~0.3$\times$ \\
MEAL V2 w/ $\heartsuit$FKD &  ResNet-50  &  25.6M & 80.91 &  95.39 & \bf ~~~0.5$\times$ \\ \hline
MEAL V2~\cite{shen2020meal}  &  MobileNet V3-S0.75 & 2.04M & 67.60 & 87.23 &  1.0 \\
MEAL V2 w/ $\heartsuit$FKD  &  MobileNet V3-S0.75 & 2.04M &  67.83 & 87.35 & \bf ~~~0.4$\times$ \\
MEAL V2~\cite{shen2020meal} &  MobileNet V3-S1.0 & 2.54M & 69.65 & 88.71 & 1.0 \\
MEAL V2 w/ $\heartsuit$FKD  &  MobileNet V3-S1.0 & 2.54M & 69.94 & 88.82 & \bf ~~~0.4$\times$ \\
\bottomrule[1.pt]
\end{tabular}
}
\vspace{-0.08in}
\end{table}

\begin{table*}[t]
\centering
\caption{FKD with supervised Vision Transformer~\cite{dosovitskiy2020image} variants (224$\times$224 input size) on ImageNet-1K. Models are trained from scratch.}
\label{tab:my-table_dataset_vit}
\vspace{-0.1in}
\resizebox{0.76\textwidth}{!}{
\begin{tabular}{l|c|c|c|c|c|c|c}
\toprule[1.0pt]
 Method &  Network   &  Epochs  &  \#Params (M) &  FLOPs (B) &  Extra Data Aug. &  Top-1 (\%) &  Speedup \\ \midrule 
DeiT~\cite{touvron2021training} w/o KD &  ViT-T   & 300  & 5.7 & 1.3 & MixUp+CutMix+RA & 72.2 & -- \\
DeiT~\cite{touvron2021training} w/ KD &  ViT-T   & 300 & 5.7 & 1.3 & MixUp+CutMix+RA & 74.5 & 1.0\\ \hline
ViT~\cite{dosovitskiy2020image} (Vanilla) &  ViT-T & 300 & 5.7  &  1.3  & \bf None & ~~~~~~~~68.7~\cite{heo2021pit} & --  \\ 
\bf ViT w/ FKD (Ours)  &  ViT-T   & 300 &  5.7 & 1.3 & \bf None & \bf 75.2 &  ~~~\bf 0.15$\times$ \\ \midrule[.8pt]
SReT~\cite{shen2021sliced} w/o KD &  SReT-LT   & 300  &  5.0   &  1.2  & MixUp+CutMix+RA & 76.7 & -- \\
SReT~\cite{shen2021sliced} w/ KD & SReT-LT    & 300  &  5.0   &  1.2  & MixUp+CutMix+RA & 77.7  &  1.0 \\ \hline
SReT~\cite{shen2021sliced} (Vanilla) &  SReT-LT  & 300  &  5.0   &  1.2  & \bf None & -- & --  \\ 
\bf SReT w/ FKD (Ours)  &  SReT-LT    & 300  &  5.0   &  1.2  & \bf None & \bf 78.7 &  ~~~\bf 0.14$\times$ \\
\bottomrule[1.0pt]
\end{tabular}
}

\vspace{-0.08in}

\centering
\caption{Ablation results (Top-1) on ImageNet-1K of different label quantization strategies. $m=8$ is used in this ablation.}
\label{tab:my-table_ablation_compression}
\resizebox{0.92\textwidth}{!}{
\begin{tabular}{l|c|c||c|c|c|c|c}
\toprule[1.pt]
 Method &  Network &  Full   &   Hard &  Smoothing &  Marginal Re-Norm ($K$=5) &  Marginal Smoothing ($K$=5)  & Marginal Smoothing ($K$=10)\\ \midrule 
MEAL V2 w/ FKD &  ResNet-50 & \bf 80.65 & 80.20 & 80.23 & 80.40 &  80.58 & 80.52 \\  
FKD (from scratch) &  ResNet-50   & 79.48  &  79.09  & 79.37 & 79.23 & \bf 79.51 & 79.44\\
\bottomrule[1.pt]
\end{tabular}
}

\vspace{-0.08in}

\centering
\caption{Ablation results (Top-1) on ImageNet-1K of different numbers ($m$) of cropping regions within an image used in a {\em mini}-batch.}
\label{tab:my-table_ablation_crops}
\resizebox{0.6\textwidth}{!}{
\begin{tabular}{l|c|c|c|c|c|c|c}
\toprule[1.pt]
 Method &  Network & \bf $m=1$   & \bf $m=2$  & \bf $m=4$  & \bf $m=8$ & \bf $m=16$   & \bf $m=32$ \\ \midrule 
Vanilla &  ResNet-50 & 77.18 & 77.91 & \bf 78.14  & 77.89 & 75.89 & 70.09\\  
MEAL V2 w/ FKD &  ResNet-50 & 80.67 & \bf 80.70 & 80.66 & 80.58 & 80.36 & 80.17\\  
FKD (from scratch) &  ResNet-50   & 79.59 & 79.62  & \bf 79.76 & 79.51 & 78.12 & 74.61\\
\bottomrule[1.pt]
\end{tabular}
}
\vspace{-0.15in}
\end{table*}

\noindent{\textbf{Baseline Knowledge Distillation Methods.}}
\vspace{-0.08in} 
\begin{itemize}[leftmargin=0.2in]
	\addtolength{\itemsep}{-0.08in}
	\item[$\blacktriangleright$] ReLabel~\cite{yun2021re} (Label Map Distillation). ReLabel used the pre-generated global label maps from the pre-trained teacher for reducing the cost on the teacher branch when conducting distillation.
	\item[$\blacktriangleright$] MEAL V2~\cite{shen2020meal} (Fine-tuning Distillation). MEAL V2 proposed to distill student network from the pre-trained parameters\footnote{The pre-trained parameter is from \href{https://github.com/rwightman/pytorch-image-models}{timm}~\cite{rw2019timm} with version$<=$0.4.12.} and giant teacher ensemble for fast convergence and better accuracy.
	\item[$\blacktriangleright$] FunMatch~\cite{yun2021re} (Oracle Distillation). FunMatch is a standard knowledge distillation framework with strong teacher models and augmentations. We consider it as the strong baseline approach for efficient KD when using the same or similar teacher supervisors.
	\item[$\vartriangleright$] S$^2$-BNN~\cite{yun2021re} (Self-supervised Distillation). A distillation solution for self-supervised learning task. The teacher is pre-learned from the self-supervised learning methods, such as MoCo V2~\cite{chen2020improved}, SwAV~\cite{caron2020unsupervised}, etc.
\end{itemize}

\subsection{Supervised Learning}

\noindent{\textbf{CNNs.}}

{\bf (i) ReLabel}~\cite{yun2021re}. The comparison with ReLabel is shown in Table~\ref{tab:my-table_relabel}, using the training settings introduced in our Appendix, which is nearly the same as ReLabel, our accuracies on ResNet-50/101 both outperform ReLabel by {\bf $\sim$1\%} with slightly faster training speed. These significant and consistent improvements of FKD show great potential for practical usages in real-world applications.

{\bf (ii) MEAL V2}~\cite{shen2020meal}. We use FKD framework to train the MEAL V2 models. The results are shown in Tabel~\ref{tab:my-table_MEALV2}. When employing the same hyper-parameters and teacher networks, FKD can speed up {\bf 2$\sim$4$\times$} for MEAL V2 {\bf without compromising accuracy}. Using {\em cosine lr} and more epochs in training further improves the accuracy.

{\bf (iii) FunMatch}~\cite{beyer2021knowledge} (Oracle). We consider FunMatch as the oracle/strong KD baseline, our plain FKD achieves 79.8\% w/o extra augmentations, which is 0.7\% lower than FunMatch (80.5\%). The result of FKD with more complex optimizers~\cite{loshchilov2018decoupled,you2017large} and more augmentations~\cite{zhang2018mixup,yun2019cutmix} (similar to FunMatch) will be presented after being explored.

\vspace{0.02in}
\noindent{\textbf{Vision Transformers.}}
\vspace{0.02in}

{\bf (i) ViT/DeiT}~\cite{dosovitskiy2020image,touvron2021training}. The results are shown in Tabel~\ref{tab:my-table_dataset_vit} of first group. Our non-extra augmentation result (75.2\%) using ViT-T backbone is better than DeiT-T with distillation (74.5\%), while we only require {\bf 0.15$\times$} resources than DeiT distillation protocol in training.

{\bf (ii) SReT}~\cite{shen2021sliced}. We also conduct our FKD using SReT-LT, our result (78.7\%) is consistently better than its original KD design (77.7\%) with fewer training resources.

\noindent{\textbf{Ablations: (i) Effects of Crop Number in Each Image.}} We explore the effect of different numbers of crops sampled from the same image within a {\em mini}-batch to the final performance. For the conventional data preparation strategy, on each image we solely sample one crop for a {\em mini}-batch to train the model. Here, we evaluate the $m$ from 1 crop to 32 crops as shown in Table~\ref{tab:my-table_ablation_crops}. Surprisingly, using a few crops from the same image leads to better performance than the single crop solution with a non-negligible margin, especially on the traditional image classification system. This indicates that the internal diversity of samples in a {\em mini}-batch has a limit for tolerance, properly reducing such diversity can boost the accuracy, while we can also observe that after $m$$>$8, the performance decreases substantially, thus the diversity is basically still critical for learning good status of the model. Nevertheless, this is a good observation for us to speedup data loading in our FKD framework.

\begin{table*}[t]
\centering
\caption{Linear evaluation results of FKD with self-supervised Binary CNN (ReActNet~\cite{liu2020reactnet}), Real-valued CNN (ResNet-50~\cite{he2016deep}).}
\label{tab:my-table_SSL_ours_com}
\vspace{-0.1in}
\resizebox{0.86\textwidth}{!}{
\begin{tabular}{l|c|c|c|c|c|c}
\toprule[1.0pt]
 Method &  Network &  Teacher &  \#Dim for Distilling & Training Epochs &  Top-1 (\%) &  Speedup \\ \midrule 
\multicolumn{6}{l}{\em Pure Distillation Based Scheme (FKD can speedup training by more than 3$\times$ with the same (similar) performance.) }\\ \hline
S$^2$-BNN~\cite{shen2021s2} &   ReActNet & MoCo V2-800ep & 128 & 200 & 61.5 &  1.0 \\ 
\bf FKD &  ReActNet & MoCo V2-800ep & 128 & 200 & 61.7 & ~~~~\bf 0.4$\times$ \\ 
S$^2$-BNN~\cite{shen2021s2} &   ResNet-50 & SwAV/RN50-w4 & 3000 & 100 & 68.7 &  1.0 \\ 
\bf FKD &  ResNet-50 & SwAV/RN50-w4 & 3000 & 100 & 68.8 & ~~~~\bf 0.3$\times$ \\ 
\bottomrule[1.0pt]
\end{tabular}
}
\vspace{-0.1in}
\end{table*}

\textbf{(ii) Different Label Compression Strategies.} We evaluate the performance for different label compression strategies. We use $m$=8 for this ablation and the results are shown in Tabel~\ref{tab:my-table_ablation_compression}. On MEAL V2 w/ FKD, we obtain the highest accuracy 80.65\% when using the full soft labels, while on the pure FKD, the best performance is from {\em Marginal Smoothing (K=5)} with 79.51\%. Increasing $K$ both decrease the accuracies in these two scenarios, we conjecture that larger $K$ will involve more noise or unnecessary minor information on the soft labels. While, they are still better than the {\em Hard} and {\em Smoothing} strategies.

\subsection{Self-Supervised Learning}

S$^2$-BNN~\cite{shen2021s2} is a pure distillation-based framework for self-supervised learning, thus the proposed FKD approach is sufficient to train S$^2$-BNN~\cite{shen2021s2} smoothly and more efficiently. We employ  SwAV~\cite{caron2020unsupervised} and MoCo V2~\cite{chen2020improved} pre-trained models as the teacher networks. Consider the more flattening distribution from the SSL learned teachers than the supervised teacher predictions (indicating that the subordinate classes from SSL trained teachers also carry crucial information), we still use the full soft label for now, and label compression strategies on SSL task will be verified further. We employ ReActNet~\cite{liu2020reactnet} and ResNet-50~\cite{he2016deep} as the target/student backbones in these experiments. The results are shown in Table~\ref{tab:my-table_SSL_ours_com}, our FKD trained models achieve slightly better performance than S$^2$-BNN with roughly 3$\times$ acceleration in training since we only use a single branch of network and no explicitly teacher forwarding existing. The slight boosts are from our liter data augmentation for FKD  when producing SSL soft labels. This is interesting in the FKD equipped SSL methods, i.e., data augmentation strategies for distillation-based SSL, and is worth exploring further. 

\subsection{Transfer Learning}

Here, we further examine whether the FKD obtained improvements on ImageNet-1K can be transferred to various downstream tasks. As shown in Tabel~\ref{tab:grid_transfer}, we present the results of object detection and instance segmentation tasks on COCO dataset~\cite{lin2014microsoft} with models pretrained on ImageNet-1K with FKD. We also employ Faster RCNN~\cite{ren2015faster} and Mask RCNN~\cite{he2017mask} with FPN~\cite{lin2017feature} following ReLabel~\cite{yun2021re}. Over the regular baseline and ReLabel, our FKD pre-trained parameters show constant gains on the downstream tasks.

\begin{table}[]
\centering
\caption{Comparison of transfer learning performance with ReLabel~\cite{yun2021re} on detection and instance segmentation tasks. The training and evaluation are conducted on COCO dataset~\cite{lin2014microsoft}.}
\label{tab:grid_transfer}
\vspace{-0.1in}
\resizebox{0.48\textwidth}{!}{
\begin{tabular}{lc|c|cc}
\toprule[1.pt]
\multirow{2}{*}{Method} & \multirow{2}{*}{Network}& \multicolumn{1} {c|} {Faster RCNN w/ FPN} & \multicolumn{2} {c} {Mask-RCNN w/ FPN}                \\ \cline{3-5} 
   &    &    bbox AP        &  bbox AP       &      mask AP    \\ \midrule
Regular Baseline &    ResNet-50  &  37.7  &    38.5   &  34.7   \\
ReLabel &  ResNet-50     & 38.2 &   39.1  &   35.2      \\
\bf FKD &   ResNet-50    &  38.7 &   39.7    &  35.9      \\
\bottomrule[1.pt] 
\end{tabular}
}
\vspace{-0.08in}
\end{table}

\begin{figure}[t]
  \centering
  \includegraphics[width=0.43\textwidth]{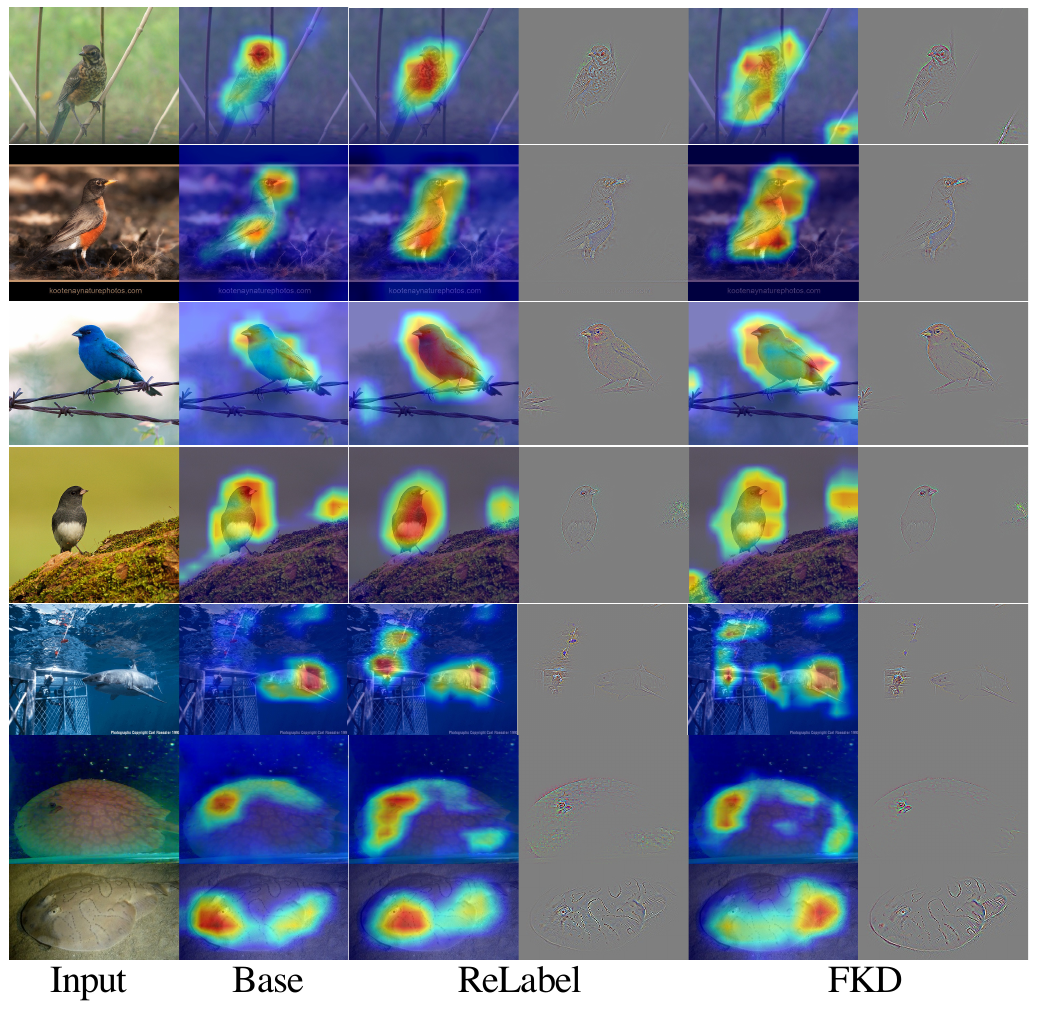}
  \vspace{-0.15in}
  \caption{Visualizations of learned attention map using GradCAM~\cite{selvaraju2017grad,jacobgilpytorchcam}. ``Base'' indicates the pre-trained PyTorch model. In each group of ReLabel and FKD, left is {\em Grad-CAM} and right is {\em Guided Backprop}.} 
  \label{fig:dis_visualization}
  \vspace{-0.15in}
\end{figure}

\subsection{Visualization, Analysis and Discussion}

To investigate the learned differences of information between ReLabel and FKD, we depict the intermediate attention maps using gradient-based localization~\cite{selvaraju2017grad}. There are three important observations that align our aforementioned analyses in Fig.~\ref{fig:dis_visualization}.

{\bf (i)} FKD's predictions are less confident than ReLabel with more surrounding context; This is reasonable since in random-crop training, many crops are basically backgrounds (context), the soft predicted label from the teacher model might be completely different from the ground-truth one-hot label and the training mechanism of FKD can leverage the additional information from context.

{\bf (ii)} FKD's attention maps have a larger active area on the object regions, which indicates that FKD trained model utilizes more cues for prediction and also captures more subtle and fine-grained information. However, it is interesting to see that the {\em guided backprop} is more focusing than ReLabel.

{\bf (iii)} ReLabel's attention is more aligned with PyTorch pre-trained model, while FKD's results are substantially unique to them. It implies that FKD's learned attention differs significantly from one-hot and global label map learned models.

\section{Conclusion}

Given its widespread use and superior performance in training compact and efficient networks, it is worthwhile investigating ways to increase the efficiency and speed of vanilla knowledge distillation. In this paper, we have presented a fast distillation framework through the pre-generated region-level soft label scheme. We have elaborately discussed the strategies of compressing soft label for practical storage and their performance comparison. We identified the observation that the input training samples within a {\em mini}-batch can be sampled from the same input image to facilitate the overhead of data loading process. We exhibit the effectiveness and adaptability of our framework by demonstrating it on supervised image classification and self-supervised representation learning tasks.

{\small
	\bibliographystyle{ieee_fullname}
	\bibliography{egbib}
}

\newpage

\appendix

\section*{\Large{Appendix}}
\vspace{1ex}

\section{Training Details and Experimental Settings}

\noindent{\textbf{Training details used in Table~\ref{tab:my-table_relabel} of the main text.}} When comparing our FKD with ReLabel~\cite{yun2021re} (Table~\ref{tab:my-table_relabel} of the main text), we use the training settings and hyper-parameters following Table~\ref{tab:my-table_relabel_ours_com}, which is nearly the same as ReLabel~\cite{yun2021re} while without {\em Warmup} and {\em Color jittering}.

\begin{table}[h]
\centering
 \vspace{-0.05in}
    \caption{Training hyper-parameters and details between ReLabel~\cite{yun2021re} and FKD used for the comparison in Table~\ref{tab:my-table_relabel} of main text.} 
    \label{tab:my-table_relabel_ours_com}
    \vspace{-0.1in}
    \resizebox{.48\textwidth}{!}{%
    \begin{tabular}{lcc}
    \toprule[1.1pt]
    Method          & ReLabel~\cite{yun2021re}    & FKD         \\
    Teacher           & EfficientNet-L2-ns-475     & EfficientNet-L2-ns-475         \\
    Epoch           & 300     & 300         \\
    Batch size      & 1,024    & 1,024         \\
    Optimizer       & SGD   & SGD       \\
    Init. {\em lr}   & 0.1   & 0.1        \\
    {\em lr} scheduler   & cosine   & cosine        \\
    Weight decay &  1e-4  &  1e-4        \\
    Random crop  & Yes   & Yes        \\
    Flipping & Yes   & Yes        \\
    Warmup epochs   & \color{red} 5       &  \color{red}0           \\
    \bf Color jittering& \bf \color{red} Yes  & \bf \color{red} No        \\
    \bottomrule[1.1pt] 
    \end{tabular}
    }
\end{table}

\noindent{\textbf{Training details used in Table 5 of the main text.}} When comparing our FKD with ViT~\cite{dosovitskiy2020image}/DeiT~\cite{touvron2021training}/SReT~\cite{shen2021sliced} (Table 5 of the main text), we use the training settings and hyper-parameters following Table~\ref{tab:my-table_distillation}.

\begin{table}[h]
\centering
    \centering
    \caption{Training hyper-parameters and details for the comparison in Table~5 of the main text when employing ViT~\cite{dosovitskiy2020image} and its variants as the backbone networks. Table is adapted from~\cite{touvron2021training}.}
    \label{tab:my-table_distillation}
    \vspace{-0.1in}
    \resizebox{.45\textwidth}{!}{%
    \begin{tabular}{lccc}
    \toprule[1.1pt]
    Method       &  ViT-B~\cite{dosovitskiy2020image}  & DeiT~\cite{touvron2021training}/SReT~\cite{shen2021sliced} &  FKD        \\
    Epoch         & 300  & 300     & 300         \\
    Batch size    & 4096 & 1024    & 1024         \\
    Optimizer     & AdamW & AdamW   & AdamW       \\
    Init. {\em lr}  & 0.003 & 0.001   & 0.002        \\
    {\em lr} scheduler   & cosine   & cosine  & cosine       \\
    Weight decay &  0.3  &  0.05  &  0.05        \\
    Warmup epochs   & 3.4       &  5  &  5          \\ \hline
    Label smoothing    &  None   &  0.1  &  None          \\
    Dropout   &   0.1 &  None &  None        \\
    Stoch. Depth   & None   &  0.1 &  0.1         \\
    Repeated Aug   & None    &  Yes & None        \\
    Gradient Clip.  & Yes    &  None & None        \\
    Rand Augment   & None     &  9/0.5 &  None         \\
    Mixup prob.   & None     & 0.8 &  None        \\
    Cutmix prob.   & None   &  1.0 &  None          \\
    Erasing prob.   & None   &  0.25  &  None       \\
    \bottomrule[1.1pt] 
    \end{tabular}
    }
\end{table}

\section{More Comparison and Results on ImageNet ReaL~\cite{beyer2020we} and ImageNetV2~\cite{recht2019imagenet} Datasets}

In this section, we provide more results on  ImageNet ReaL~\cite{beyer2020we} and ImageNetV2~\cite{recht2019imagenet} datasets. On ImageNetV2~\cite{recht2019imagenet}, we verify our FKD models on three metrics ``Top-Images'', ``Matched Frequency'', and ``Threshold
0.7'' as ReLabel~\cite{yun2021re}. We conduct experiments on two network structures: ResNet-50 and ResNet-101. The results are shown in Table~\ref{tab:my-table_INV2}, we achieve consistent improvement over baseline ReLabel on both ResNet-50 and ResNet-101.

\begin{table}[h]
\centering
\caption{Results of FKD on ImageNet ReaL~\cite{beyer2020we} and ImageNetV2~\cite{recht2019imagenet} with ResNet-\{50, 101\}. $^*$ indicates that the results are tested using their provided pre-trained model.}
\label{tab:my-table_INV2}
\vspace{-0.1in}
{\renewcommand{\arraystretch}{0.9}
\resizebox{.48\textwidth}{!}{%
\begin{tabular}{lccccc}
\toprule[1.1pt]
Method  & ImageNet & ReaL & ImageNetV2  & ImageNetV2 & ImageNetV2   \\ 
        &      &    & Top-images & Matched-frequency & Threshold-0.7 \\ \hline
\multicolumn{2}{l}{{\em ResNet-50}: }\\
ReLabel~\cite{yun2021re} &   78.9  & 85.0 &  80.5 & 67.3 &  76.0   \\ 
\bf FKD     &   \bf 79.8  &  \bf 85.5  & \bf 81.0  &  \bf 68.1  &  \bf  76.7  \\ \hline
{\em ResNet-101}: \\
ReLabel~\cite{yun2021re}$^*$ &   80.7  & 86.5  &  82.4 & 69.7 &  78.2   \\ 
\bf FKD     &   \bf 81.7   &  \bf 87.0   & \bf 83.1   &  \bf 70.5  &  \bf  78.9  \\ 
\bottomrule[1.1pt] 
\end{tabular}
}
}
\end{table}

\begin{figure}[h]
  \centering
  \includegraphics[width=0.45\textwidth]{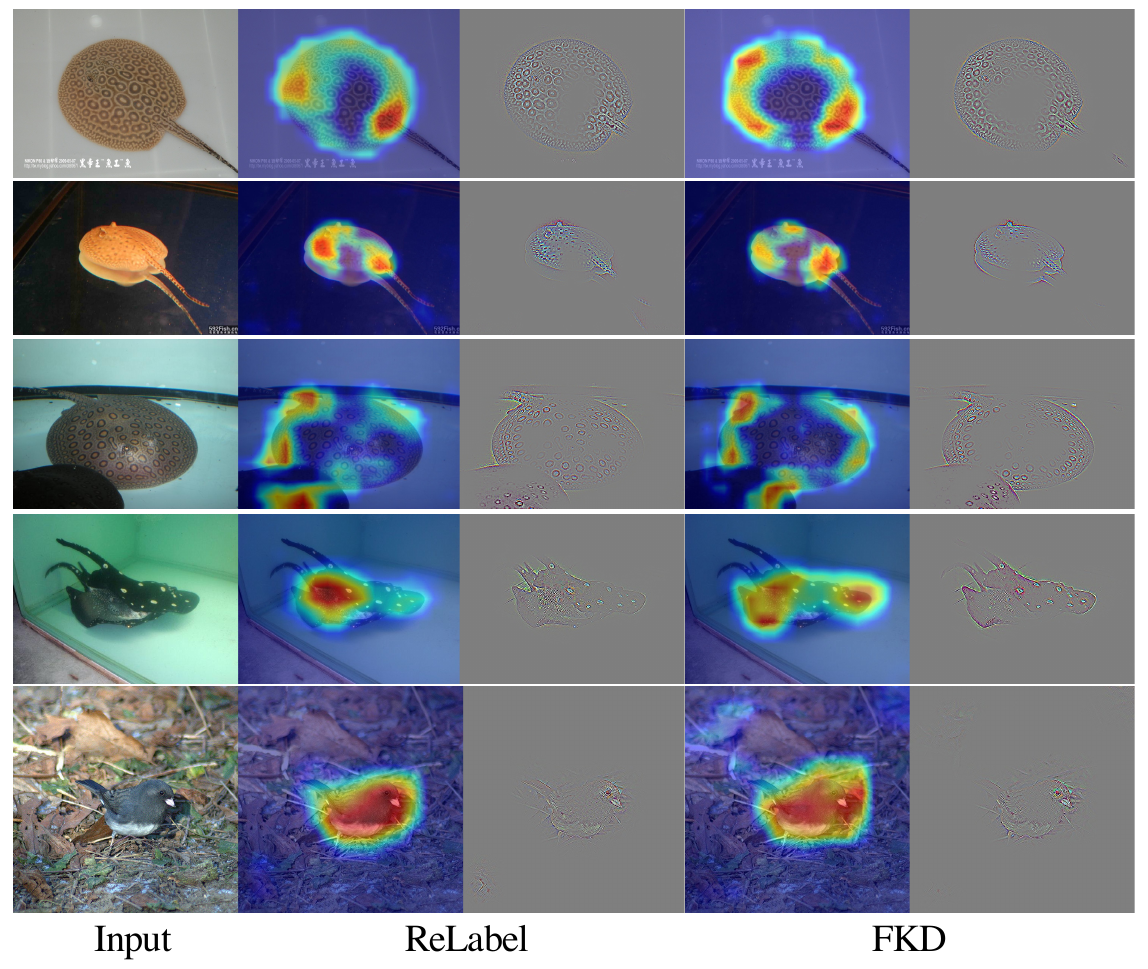}
  \vspace{-0.15in}
  \caption{More visualizations of response/attention maps.} 
  \label{fig:more_att_visualization}
  \vspace{-0.1in}
\end{figure}

\section{More Visualizations}
We provide more visualizations of the intermediate attention maps from ResNet-50 to explore the learned differences of information between ReLabel and FKD, as shown in Fig.~\ref{fig:more_att_visualization}. The observation is consistent to our main text.

\end{document}